# DRIMV_TSK: An Interpretable Surgical Evaluation Model for Incomplete Multi-View Rectal Cancer Data

Wei Zhang,[1,2] Zi Wang,[3] Hanwen Zhou,[2] Zhaohong Deng,[2] Weiping Ding,[1] Yuxi Ge,[3] Te Zhang,[4] Yuanpeng Zhang,[5] Kup-Sze Choi,[6] Shitong Wang,[2] Shudong Hu,[3]

[1] School of Artificial Intelligence and Computer Science, Nantong University, Nantong, Jiangsu, 226019, China
[2] School of Artificial Intelligence and Computer Science, Jiangnan University and Jiangsu Key Laboratory of Digi-tal Design and Software Technology, Wuxi 214122, China
[3] Department of Radiology, Affiliated Hospital of Jiangnan University, Wuxi 214000, China
[4] School of Computer Science, University of Nottingham, Nottingham, NG8 1BB, UK
[5] The Department of medical informatics, Nantong University
[6] Centre for Smart Health, Hong Kong Polytechnic University, Hong Kong
dengzhaohong@jiangnan.edu.cn

**Abstract.** A reliable evaluation of surgical difficulty can improve the success of the treatment for rectal cancer and the current evaluation method is based on clinical data. However, more data about rectal cancer can be collected with the development of technology. Meanwhile, with the development of artificial intelligence, its application in rectal cancer treatment is becoming possible. In this paper, a multi-view rectal cancer dataset is first constructed to give a more comprehensive view of patients, including the high-resolution MRI image view, pressed-fat MRI image view, and clinical data view. Then, an interpretable incomplete multi-view surgical evaluation model is proposed, considering that it is hard to obtain extensive and complete patient data in real application scenarios. Specifically, a dual representation incomplete multi-view learning model is first proposed to extract the common information between views and specific information in each view. In this model, the missing view imputation is integrated into representation learning, and second-order similarity constraint is also introduced to improve the cooperative learning between these two parts. Then, based on the imputed multi-view data and the learned dual representation, a multi-view surgical evaluation model with the TSK fuzzy system is proposed. In the proposed model, a cooperative learning mechanism is constructed to explore the consistent information between views, and Shannon entropy is also introduced to adapt the view weight. On the MVRC dataset, we compared it with several advanced algorithms and DRIMV_TSK obtained the best results.

## 1 Introduction

Currently, rectal cancer is one of the major gastrointestinal cancers, which is a serious threat to human health. Meanwhile, with the development of artificial intelligence, some artificial intelligence models have been successfully used in rectal cancer diagnosis and treatment and have received great attention. For example, Shi et al. use radiomics and deep convolutional networks to predict the effectiveness of radiotherapy for advanced rectal cancer patients [1]. Similarly, Shayesteh et al. predicted the effectiveness of radiotherapy for patients by constructing an intelligent model to explore rectal cancer MRI images [2]. Based on multi-instance learning with multi-scale convolutional neural networks, Zhang et al. explored rectal cancer MRI images and predicted the effectiveness of radiotherapy for patients [3]. Since a part of rectal cancer patients present with synchronous metastasis, the survival of these patients varies greatly, for which Zhao et al. constructed a survival prediction model for rectal cancer using logistic regression and Cox proportional risk model [4]. These methods mainly use machine learning methods to predict the effect of radiotherapy or to predict the survival of patients. In currently, total mesorectal excision (TME) is the main treatment for rectal cancer, in which the difficulty and risk of the surgery vary between different patients. Therefore, if a reliable assessment of the surgery can be conducted with the help of artificial intelligence would greatly improve the success of the procedure and have a positive impact on postoperative recovery.

The current assessment method of surgical difficulty mainly relies on the physician's judgment based on the clinical data (such as gender, age, pelvic size, and depth) [5, 6], which is a relatively subjective approach. Meanwhile, owing to the difference in the physical condition of different patients, radiotherapy or chemotherapy performed before surgery may lead to significant edema or inflammation in the rectal wall and rectal mesenteric fascia in some patients or insufficient tumor regression in some patients, all of which will increase the uncertainty of surgery, and this uncertain information cannot be represented by clinical data [7, 8]. In addition, besides the clinical data, we can collect more and more data with the development of technology. For example, multiple

MRI examinations of patients are required before surgery, so MRI images of patients from different periods and types, as well as traditional clinical data can be considered pathological data from different views. These data can be used for subsequent surgical difficulty assessment. Therefore, if we can use this multi-view pathological data and construct an artificial intelligence model to accurately assess the difficulty of surgery will greatly reduce the burden of doctors and improve the success rate of surgery.

In the area of artificial intelligence, multi-view/multi-modal learning has received great attention in the last decade, which aims to mine data with the help of intelligent models and assist people in making decisions. The representative methods are as follows: Chao et al. proposed the multi-view maximum entropy discrimination (AMVMED) algorithm for classification with the help of the marginal consistency strategy [9]. Exploring the visual, semantic, and view consistency information, Zhang et al. proposed a multi-view classification model [10]. Yang et al. proposed an adaptive multi-view fusion learning method that fuses the original data into a discriminative hidden representation for classification [11]. Xu et al. proposed a deep multi-view classification method, which explores the specific information of each view and interaction information between views for classification [12]. Although these methods have obtained excellent performance, they still face two main issues. First, these methods can be constructed on complete multi-view data, but it is difficult to collect complete data in real applications. For example, we are very difficult to record all patients' test data in a medical scenario, while for some tests not all patients are performed. Therefore, we should research new strategies to address this challenge. The second challenge is that most methods only focus on the performance of the model but ignore the interpretability of the model. This leads to most methods cannot be deployed in scenarios with high interpretability requirements, such as in healthcare.

For the first challenge, incomplete multi-view learning has achieved significant interest, with the primary strategy being the extraction of a common representation across views. For instance, Zhao et al explored a common representation from incomplete multi-view data by integrating Non-Negative Matrix Factorization (NMF) and manifold regularization for modeling [13]. Similarly, Wen et al also introduced NMF to explore common representation and used graph regularization to enhance the discriminability of common representation [14]. Based on self-representation learning and tensor Schatten p-norm constraint, Lv et al. learned a common representation for modeling [15]. Zhou et al. first learned similarity graphs of each incomplete view, and then merged them into a common graph for modeling [16]. There is no doubt that the above methods are effective, but they still ignore an important point that is there not only common information between incomplete multi-view data but also specific information in each view. If both two types of information can be explored and combined to construct a model, this challenge will be addressed more effectively.

For the second challenge, the interpretability of intelligent models is crucial in some sensitive scenarios, especially medical scenarios. TSK fuzzy system is a rule-based model known for its powerful data-driven learning capability [17], and it has been applied in some multi-view medical scenarios. For instance, Jiang et al. introduced a TSK fuzzy system with the help of cooperative learning for multi-view epilepsy data detection [18]. Similarly, In order to further enhance the reliability of epilepsy detection, Tian et al. first introduced deep learning to extract the multi-view epilepsy feature and then designed a multi-view TSK fuzzy system for modeling [19]. In addition, based on the TSK fuzzy system, a new multi-view transfer method is constructed to deal with the problem of insufficient labeled data in epilepsy detection [20]. Nevertheless, these available multi-view TSK fuzzy systems mostly focus on brain diseases, while there is no reliable fuzzy system for rectal diseases. Moreover, these methods can only handle complete multi-view data. Therefore, a new effective TSK fuzzy system for incomplete multi-view data is needed.

To resolve the above-mentioned issues, the attempt of interpretable surgical difficulty assessment is conducted in this paper. First, to the best of our knowledge, there is no published rectal cancer dataset for the surgical difficulty assessment. In this paper, a multi-view rectal cancer surgery difficulty assessment dataset is constructed (MVCR) jointly with the Affiliated Hospital of Jiangnan University. MVCR includes three views, i.e., high-resolution and pressed-fat MRI image views, as well as a traditional clinical data view. Among them, the high-resolution MRI image has high soft tissue resolution, which can accurately show the relationship between the tumor and the layers of the rectal intestinal wall, and also clearly shows the adjacent anatomical structures such as the rectal mesenteric fascia. The pressed fat MRI image can clearly show both the tumor symbol and the pelvic anatomy, and it can also clearly show the pelvic edema signal. Combing these data, we can have a comprehensive overview of the patient's rectal lesion and the tissue within the rectal mesentery and it can help us to achieve a more accurate assessment of the difficulty of the procedure.

Then, a novel incomplete multi-view fuzzy system is proposed to address incomplete data scenarios in medical applications. Specifically, a novel dual representation learning method with the help of matrix factorization for incomplete multi-view data is proposed first. In the proposed method, both the common information between views and specific information in each view are extracted, and the missing view imputation is also unified with the representation learning process. In this way, missing view imputation and dual representation learning can be negative to each other. Moreover, a second-order similar learning mechanism is further introduced to enhance the cooperation between missing view imputation and representation learning. Finally, an interpretable multi-view TSK fuzzy system with dual hidden view cooperative learning is proposed for surgical difficulty assessment. In this method, not only the imputed multi-view rectal cancer data but also the learned common and specific hidden views are combined to construct the assessment model. In addition, a cooperative learning mechanism is introduced to explore the consistent information between views, and the negative Shannon entropy is further constructed to learn the optimal view weight. Based on these two mechanisms, we can explore the rectal cancer data fully and provide the optimal surgical difficulty assessment.

The main contributions are outlined below:

1) To the best of our knowledge, the first rectal cancer multi-view dataset for surgical difficulty assessment is constructed.
2) To the best of our knowledge, the first interpretable incomplete multi-view TSK fuzzy system is proposed for rectal cancer surgical difficulty assessment. In this method, we first construct a dual representation learning method, which integrates the missing view imputation and dual hidden representation learning. Then, we integrate imputed data and dual hidden views to construct the surgical difficulty assessment model.
3) The proposed method is validated through extensive experiments, and its advantages are analyzed in depth.

The remainder of this paper is organized as follows. Section II provides a brief overview of background. Section III details the proposed method. Experimental studies conducted on various datasets are presented in Section IV. Finally, Section V concludes this paper and outlines future works.

## 2   Background

### 2.1   Takagi-Sugeno-Kang Fuzzy System

TSK fuzzy system is a model that utilizes fuzzy sets and fuzzy logic to perform intelligent reasoning [21], which not only has powerful data-driven abilities but also balances performance and interpretability compared to traditional machine learning models [17]. The fuzzy rule of the TSK fuzzy system is shown below:

$$R^k: IF\ x_1\ is\ A_1^k\ \wedge \cdots \wedge\ x_d\ is\ A_d^k$$
$$THEN\ f_k^1(x) = p_0^{k,1} + p_1^{k,1}x_1 + \cdots + p_d^{k,1}x_d,$$
$$\vdots$$
$$f_k^c(x) = p_0^{k,c} + p_1^{k,c}x_1 + \cdots + p_d^{k,c}x_d,$$
$$\vdots$$
$$f_k^C(x) = p_0^{k,C} + p_1^{k,C}x_1 + \cdots + p_d^{k,C}x_d,$$
$$k = 1,2,\ldots,K, c = 1,2,\ldots,C \tag{1}$$

where $\mathbf{x} = [x_1, x_2, \ldots, x_d]$ is the input vector, d is the feature dimension, $\wedge$ is the fuzzy conjunction operator, $K$ is the number of rules, and $C$ is the output dimension. $A_j^k$ is a fuzzy subset of the $k$-th rule associated with $j$-th feature, and the fuzzy subset consists of the membership function, where $1 \leq j \leq d$. Different membership functions for fuzzy sets can be chosen according to different application scenarios. The Gaussian function is used as a membership function in the proposed method [22], and it is given as following:

$$\mu_{A_j^k}(x_j) = exp\left(-(x_j - e_{k,j})^2 / 2q_{k,j}\right) \tag{2a}$$

where $e_{k,j}$ and $q_{k,j}$ is the antecedent parameters of the TSK fuzzy system and they represent the center and width of the membership function, respectively. These two parameters were estimated by Fuzzy C-means Clustering [23] in past papers, but a more stable method [24] is used in this paper to overcome the random initialization of FCM.

When the antecedent parameters have been estimated, the firing strength and normalized firing strength of the $k$-th rule for each input vector can be obtained as follows:

$$\mu^k(\mathbf{x}) = \prod_{j=1}^d \mu_{A_j^k}(x_j) \tag{2b}$$

$$\tilde{\mu}^k(\mathbf{x}) = \frac{\mu^k(\mathbf{x})}{\sum_{k'=1}^K \mu^{k'}(\mathbf{x})} \tag{2c}$$

The output vector of the TSK fuzzy system by combining all the rules is defined below:

$$\mathbf{y} = \sum_{k=1}^K \tilde{\mu}^k(\mathbf{x}) \mathbf{f}_k(\mathbf{x}) \tag{2d}$$

where $\mathbf{f}_k(\mathbf{x}) = [f_k^1(x), f_k^2(x), \ldots, f_k^C(x)]$ is the output of the $k$-th rule. When the $e_{k,j}$ and $q_{k,j}$ are calculated, (2d) can be reformulated as below:

$$\mathbf{y} = \mathbf{x}_g \mathbf{P}_g \in R^{1 \times C} \tag{3a}$$

The definitions of $\mathbf{x}_g$ and consequent parameters $\mathbf{P}_g$ are given below:

$$\mathbf{x}_e = [1, \mathbf{x}] \in R^{1 \times (1+d)} \tag{3b}$$

$$\tilde{\mathbf{x}}^k = \tilde{\mu}^k(\mathbf{x})\mathbf{x}_e \in R^{1 \times (1+d)} \tag{3c}$$

$$\mathbf{x}_g = [\tilde{\mathbf{x}}^1, \tilde{\mathbf{x}}^2, \ldots, \tilde{\mathbf{x}}^K] \in R^{1 \times K(1+d)} \tag{3d}$$

$$\mathbf{p}_k^c = [p_0^{k,C}, p_1^{k,C}, \ldots, p_d^{k,C}] \in R^{1 \times (1+d)} \tag{3e}$$

$$\mathbf{P}^c = [\mathbf{p}_1^c, \mathbf{p}_2^c, \ldots, \mathbf{p}_K^c] \in R^{1 \times K(1+d)} \tag{3f}$$

$$\mathbf{P}_g = [(\mathbf{P}^1)^{\mathrm{T}}, (\mathbf{P}^2)^{\mathrm{T}}, \ldots, (\mathbf{P}^C)^{\mathrm{T}}] \in R^{K(1+d) \times C} \quad (3g)$$

### 2.2 Incomplete Multi-view Learning

In many real-world applications, especially in medical scenarios, incomplete multi-view data is a common issue. Several methods have been proposed in recent years to address this challenge. A natural method uses missing data imputation methods to address this issue [25, 26]. Besides these methods, some representation learning based methods have received great attention and these methods normally extract a common hidden representation between incomplete views to address this issue. For example, On the basis of NMF, the common representation between incomplete views was extracted for modeling in [27]. To further alleviate the negative effects of missing views, Shao et al. integrated weighted NMF and $L_{2,1}$ regularization to extract common representation [28]. Similarly, Hu et al. introduced semi-NMF to explore a common representation between views [29]. In order to furthermore enhance the discriminability of the extracted complete common representation, efforts have been made to integrate it with missing view imputation. For instance, based on matrix factorization, the common representation learning and missing view imputation were unified into the unified framework in [30]. In the kernel space, Liu et al. unified common representation learning and missing view imputation into a common optimization framework by exploring the complementary information between views [31]. Similarly, Based on self-representation learning and hypergraph learning, the common representation learning and missing view imputation were integrated into the unified framework in [32].

Moreover, some grouping strategy based methods are also proposed. For instance, by partitioning the available incomplete multi-view data into multiple complete datasets for modeling, an incomplete multi-view classification method is proposed in [33]. Following the IMSF, Xiang et al. proposed an enhanced classification method by introducing sparse learning [34]. In addition, Liu et al. introduced hypergraph learning to explore the deep relationships that exist within each grouped data [35]. Similarly, Thung et al. introduced deep learning to mine the nonlinear relationship with each block of grouped data [36].

Although the above methods produced excellent successes, they still face some issues. First, these methods mostly explore the common information between views, but neglect the specific information in each view. Therefore, these methods cannot fully explore the incomplete multi-view data and make the constructed model for subsequent tasks relatively weakly robust. Second, the complementary information between views is usually ignored in grouping strategy based methods. For that purpose, a new incomplete multi-view learning method is proposed. First, a new dual representation learning for incomplete multi-view data is constructed. The proposed dual representation method not only extracts the completed common and specific representations but also unifies missing view imputation and dual representation learning. Then, an interpretable multi-view TSK fuzzy system is constructed by combining both the dual representations and imputed data.

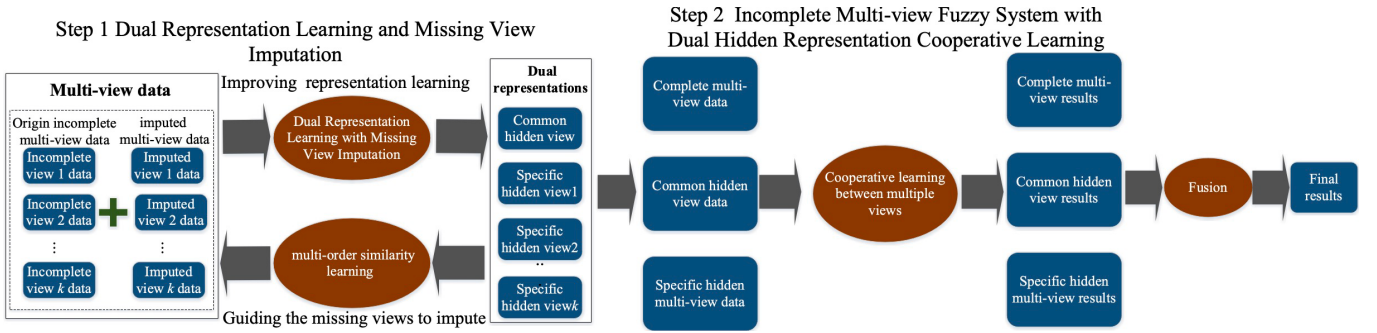

**Fig. 1.** The framework of the proposed DRIMV_TSK

## 3 INCOMPLETE MULTI-VIEW FUZZY SYSTEM

### 3.1 This The Framework of the Proposed Incomplete Multi-view Learning

In real applications, the gathered multi-view/multi-modal data may be incomplete, which is especially true in medical scenarios. Therefore, a reliable incomplete multi-view learning method is urgently required. In this paper, an incomplete multi-view learning method is proposed for rectal cancer surgical difficulty assessment, and the framework is shown in Fig. 1. The framework is composed of two main parts, the first part is dual representation learning for incomplete multi-view data, in which the common information between views and specific information in each view are explored simultaneously. Moreover, in order to further enhance the quality of the extracted representations, the missing view imputation is also introduced into the representation learning. In this way, missing view imputation and representation learning can be negative to each other. Then, the second part is the multi-view fuzzy system modeling with dual representation cooperative learning. In this part, to fully explore the incomplete multi-view data, the learned dual hidden representations and imputed visible data are integrated to build a multi-view fuzzy system for surgical difficulty assessment. Finally, A cooperative learning mechanism and adaptively view weighting mechanism are also introduced to enhance the assessment accuracy.

## 3.2 Incomplete Dual Representation Learning and Missing View Imputation

Given an incomplete multi-view dataset $\{\mathbf{X}^v \in R^{N \times d^v}, v = 1, 2, \dots, V\}$, a commonly used strategy is to learn a common representation from incomplete multi-view data. In recent years, few methods enhanced the discriminability of the learned common representation by introducing missing view imputation into it. For example, on the basis of matrix factorization, in [37], the missing view imputation was unified into common representation learning and geometric structure preservation was introduced to improve the performance of the model. The objective function is as follows:

$$\min_{\mathbf{U}^v, \mathbf{H}, \mathbf{B}^v} \sum_{v=1}^{V} \|\mathbf{W}^v(\mathbf{X}^v + \mathbf{E}^v \mathbf{U}^{v,T} - \mathbf{H}\mathbf{B}^v)\|_2 + \beta \|\mathbf{H}\|_{2,1} + \gamma \sum_{v=1}^{V} tr(\mathbf{U}^{v,T} \mathbf{L}^v \mathbf{U}^v)$$
$$s.t. \ \mathbf{H}, \mathbf{B}^v \geq 0 \tag{4}$$

where $\mathbf{W}^v \in R^{N \times N}$ is the sample weighting matrix of each view, and it also is a diagonal matrix definied as below:

$$\mathbf{W}_{i,i}^v = \begin{cases} 1, & \text{if } i\text{th instance is complete} \\ w, & \text{otherwise} \end{cases} \tag{5}$$

where $w$ is the sample weighted value and it is the ratio of the number of available instances to the total number of instances. $\mathbf{E}^v \in R^{N \times N}$ is the diagonal matrix used as an indicator. If the $i$-th instance is complete in $v$-th view, then $\mathbf{E}_{i,i}^v = 0$, otherwise $\mathbf{E}_{i,i}^v = 1$. $\mathbf{U}^v \in R^{d^v \times N}$ is the error matrix of each view and it is introduced to impute missing views. $\mathbf{H} \in R^{N \times m}$ is the common representation. $\mathbf{B}^v \in R^{m \times d^v}$ is the base matrix of $v$-th view. $m$ is the features dimension of common representation. $\mathbf{L}^v \in R^{d^v \times d^v}$ is the Laplacian matrix. (4) is constructed with three parts. This first part is NMF and is used to extract the common representation, the second part is used to enhance the robustness, and the third part is used to mine the similarity relationships within views to enhance the quality of imputed views.

It should be noted that there exists both common information and specific information in multi-view data [38]. (4) obviously ignores specific information in each view. Therefore, similar to (4), an enhanced incomplete multi-view representation learning is constructed to extract the common and specific information simultaneously. In this paper, we redefine the error matrix $\mathbf{U}^v \in R^{N \times d^v}$ and $\widetilde{\mathbf{X}}^v = \mathbf{X}^v + \mathbf{E}^v \mathbf{U}^v$, the objective function is shown as below:

$$\min_{\mathbf{U}^v, \mathbf{H}, \mathbf{B}^v} \sum_{v=1}^{V} \|\widetilde{\mathbf{X}}^v - \mathbf{H}_s^{v,T} \mathbf{B}_s^v - \mathbf{H}_c^T \mathbf{B}_c^v\|_F + \lambda_1 \sum_{v=1}^{V} \|\mathbf{H}_s^{v,T} \mathbf{H}_c\|_F \tag{6}$$

where $\mathbf{H}_s^v \in R^{m_s \times N}$ and $\mathbf{H}_c \in R^{m_c \times N}$ represents the specific and common representations. $\mathbf{B}_s^v \in R^{m_s \times d^v}$ and $\mathbf{B}_c^v \in R^{m_c \times d^v}$ are the mapping matrix for specific and common representation, respectively. $m_s$ and $m_c$ are the feature dimensions for specific and common representations, respectively. To simplify the model, we denote $m_s = m_c = m$ in this paper. In (6), the first part is used to learn common and specific representations, and the second part is the orthogonal constraint, which is used to minimize the redundant information between two representations.

In (4), Laplacian regularization is introduced to enhance the quality of imputed views. This method ignores the similar information between pairwise data instances. In addition, (4) can extract similar information within each view and ignore the consistent information between views. To this end, a second-order similar learning method is introduced to address these shortcomings, and the objective function is defined as follows:

$$\min_{\mathbf{U}^v} \sum_{v=1}^{V} \sum_{i=1}^{N} \sum_{j=1}^{N} \mathbf{G}_{s,ij}^v \|\widetilde{\mathbf{X}}_i^k - \widetilde{\mathbf{X}}_j^k\|_2^2 + \sum_{v=1}^{V} \sum_{i=1}^{N} \sum_{j=1}^{N} \mathbf{G}_{c,ij} \|\widetilde{\mathbf{X}}_i^k - \widetilde{\mathbf{X}}_j^k\|_2^2 + \sum_{v=1}^{V} \sum_{i=1}^{N} \|\widetilde{\mathbf{X}}_i^k - \sum_{j=1}^{N} \mathbf{G}_{s,ij}^v \widetilde{\mathbf{X}}_j^k\|_2^2 + \sum_{v=1}^{V} \sum_{i=1}^{N} \|\widetilde{\mathbf{X}}_i^k - \sum_{j=1}^{N} \mathbf{G}_{c,ij} \widetilde{\mathbf{X}}_j^k\|_2^2 \tag{7}$$

where $\mathbf{G}_c \in R^{N \times N}$ and $\mathbf{G}_s^v \in R^{N \times N}$ are the similar matrix of common and specific representation, respectively. $\mathbf{G}_{s,ij}^v$ is defined as follows:

$$\mathbf{G}_{s,ij}^v = \begin{cases} \Psi(\mathbf{H}_{s,i}^{v,T}, \mathbf{H}_{s,j}^{v,T}), & \text{if } \mathbf{H}_{s,i}^{v,T} \text{ is the p-nearest} \\ & \text{neighbor of } \mathbf{H}_{s,j}^{v,T} \\ 0, & \text{otherwise} \end{cases} \tag{8}$$

where $\mathbf{H}_{s,i}^{v,T}$ is the $i$-th instance associated with $v$-th specific representation, $\Psi(*,*)$ is the Gaussian function. The definition of $\mathbf{G}_{c,ij}$ is similar to $\mathbf{G}_{s,ij}^v$, and the only difference is that $\mathbf{G}_{c,ij}$ is estimated based on $\mathbf{H}_{c,i}^T$. In (7), the first two terms are the first-order similar learning mechanism, and these two terms explore the similar information between two instances within the common and specific views. The last two terms are the second-order similar learning mechanism, which explores the similar information between pairwise data instances within the common and specific views. Through (7), the second-order similar information between views and within each view can be fully explored. Based on [39], (7) can be transformed as follows:

$$\min_{\mathbf{U}^v} \sum_{v=1}^{V}\sum_{i=1}^{N}\sum_{j=1}^{N} \mathbf{G}_{s,ij}^v \left\|\widetilde{\mathbf{X}}_i^k - \widetilde{\mathbf{X}}_j^k\right\|_2^2 + \sum_{v=1}^{V}\sum_{i=1}^{N}\sum_{j=1}^{N} \mathbf{G}_{c,ij} \left\|\widetilde{\mathbf{X}}_i^k - \widetilde{\mathbf{X}}_j^k\right\|_2^2 + \sum_{v=1}^{V}\sum_{i=1}^{N} \left\|\widetilde{\mathbf{X}}_i^k - \sum_{j=1}^{N}\mathbf{G}_{s,ij}^v \widetilde{\mathbf{X}}_j^k\right\|_2^2 +$$

$$\sum_{v=1}^{V}\sum_{i=1}^{N}\left\|\widetilde{\mathbf{X}}_i^k - \sum_{j=1}^{N}\mathbf{G}_{c,ij}\widetilde{\mathbf{X}}_j^k\right\|_2^2 = \sum_{v=1}^{V} tr(\widetilde{\mathbf{X}}^{v,\mathrm{T}}\mathbf{L}_s^v\widetilde{\mathbf{X}}^v) + \sum_{v=1}^{V} tr(\widetilde{\mathbf{X}}^{v,\mathrm{T}}\mathbf{L}_c\widetilde{\mathbf{X}}^v) + \sum_{v=1}^{V} tr(\widetilde{\mathbf{X}}^{v,\mathrm{T}}\boldsymbol{\Lambda}_s^{v,\mathrm{T}}\boldsymbol{\Lambda}_s^v\widetilde{\mathbf{X}}^v)$$

$$+ \sum_{v=1}^{V} tr(\widetilde{\mathbf{X}}^{v,\mathrm{T}}\boldsymbol{\Lambda}_c^{\mathrm{T}}\boldsymbol{\Lambda}_c\widetilde{\mathbf{X}}^v) \tag{9}$$

where $\mathbf{L}_s^v = \mathbf{D}_s^v - \mathbf{G}_s^v$ and $\mathbf{L}_c = \mathbf{D}_c - \mathbf{G}_c$ are the Laplacian matrix of the common and specific representation, respectively. $\mathbf{D}_s^v$ and $\mathbf{D}_c$ are the diagonal matrix and $\mathbf{D}_{s,i,i}^v = \sum_{j=1}^{N}\mathbf{G}_{s,i,j}^v$, $\mathbf{D}_{c,i,i}^v = \sum_{j=1}^{N}\mathbf{G}_{c,i,j}^v$. Besides, $\boldsymbol{\Lambda}_s^v = \mathbf{I} - \mathbf{G}_s^v$ and $\boldsymbol{\Lambda}_c = \mathbf{I} - \mathbf{G}_c$. $\mathbf{I} \in R^{N \times N}$ is the identity matrix. Combing (6) and (9), the objective function is updated as below:

$$\min_{\mathbf{U}^v, \mathbf{H}_s^v, \mathbf{B}_s^v, \mathbf{H}_c, \mathbf{B}_c} \sum_{v=1}^{V}\left\|\widetilde{\mathbf{X}}^v - \mathbf{H}_s^{v,\mathrm{T}}\mathbf{B}_s^v - \mathbf{H}_c^{\mathrm{T}}\mathbf{B}_c^v\right\|_F + \lambda_1 \sum_{v=1}^{V}\left\|\mathbf{H}_s^{v,\mathrm{T}}\mathbf{H}_c\right\|_F + \lambda_2\left(\sum_{v=1}^{V} tr(\widetilde{\mathbf{X}}^{v,\mathrm{T}}\mathbf{L}_s^v\widetilde{\mathbf{X}}^v) + \sum_{v=1}^{V} tr(\widetilde{\mathbf{X}}^{v,\mathrm{T}}\mathbf{L}_c\widetilde{\mathbf{X}}^v)\right) +$$

$$\lambda_3\left(\sum_{v=1}^{V} tr(\widetilde{\mathbf{X}}^{v,\mathrm{T}}\boldsymbol{\Lambda}_s^{v,\mathrm{T}}\boldsymbol{\Lambda}_s^v\widetilde{\mathbf{X}}^v) + \sum_{v=1}^{V} tr(\widetilde{\mathbf{X}}^{v,\mathrm{T}}\boldsymbol{\Lambda}_c^{\mathrm{T}}\boldsymbol{\Lambda}_c\widetilde{\mathbf{X}}^v)\right) \tag{10}$$

Based on (10), the missing view imputation and dual representation learning are unified into one framework, where these two parts negotiate with each other and both the common and specific information can be extracted. $\lambda_1$, $\lambda_2$ and $\lambda_3$ are the regularization hyperparameters and they are introduced to balance the factors to the corresponding terms.

Based on [40], the iterative optimization method is used to solve (10), and the details of the solution process are shown below:

1) Update $\mathbf{U}^v$ with $\mathbf{H}_s^v, \mathbf{B}_s^v, \mathbf{H}_c, \mathbf{B}_c^v$ fixed
When fixing the other variables except $\mathbf{U}^v$, the minimization problem is defined below:

$$\min_{\mathbf{U}^v} \sum_{v=1}^{V}\left\|\widetilde{\mathbf{X}}^v - \mathbf{H}_s^{v,\mathrm{T}}\mathbf{B}_s^v - \mathbf{H}_c^{\mathrm{T}}\mathbf{B}_c^v\right\|_F + \lambda_2\left(\sum_{v=1}^{V} tr(\widetilde{\mathbf{X}}^{v,\mathrm{T}}\mathbf{L}_s^v\widetilde{\mathbf{X}}^v) + \sum_{v=1}^{V} tr(\widetilde{\mathbf{X}}^{v,\mathrm{T}}\mathbf{L}_c\widetilde{\mathbf{X}}^v)\right)$$

$$+\lambda_3\left(\sum_{v=1}^{V} tr(\widetilde{\mathbf{X}}^{v,\mathrm{T}}\boldsymbol{\Lambda}_s^{v,\mathrm{T}}\boldsymbol{\Lambda}_s^v\widetilde{\mathbf{X}}^v) + \sum_{v=1}^{V} tr(\widetilde{\mathbf{X}}^{v,\mathrm{T}}\boldsymbol{\Lambda}_c^{\mathrm{T}}\boldsymbol{\Lambda}_c\widetilde{\mathbf{X}}^v)\right) \tag{11}$$

By denoting $\widetilde{\boldsymbol{\Lambda}}_c = \boldsymbol{\Lambda}_c^{\mathrm{T}}\boldsymbol{\Lambda}_c$, $\widetilde{\boldsymbol{\Lambda}}_s^v = \boldsymbol{\Lambda}_s^{v,\mathrm{T}}\boldsymbol{\Lambda}_s^v$, and taking the derivative of (11) with respect to $\mathbf{U}^v$ and setting it to zero, the rules for updating $\mathbf{U}^v$ is given below:

$$\mathbf{U}^v = \left(\mathbf{E}^{v,\mathrm{T}}\mathbf{E}^v + \lambda_2 \mathbf{E}^{v,\mathrm{T}}(\mathbf{L}_s^v + \mathbf{L}_c)\mathbf{E}^v + \lambda_3 \mathbf{E}^{v,\mathrm{T}}(\widetilde{\boldsymbol{\Lambda}}_s^v + \widetilde{\boldsymbol{\Lambda}}_c)\mathbf{E}^v\right)^{-1}$$

$$\left(\mathbf{E}^{v,\mathrm{T}}\mathbf{H}_s^{v,\mathrm{T}}\mathbf{B}_s^v + \mathbf{E}^{v,\mathrm{T}}\mathbf{H}_c^{\mathrm{T}}\mathbf{B}_c^v - \mathbf{E}^{v,\mathrm{T}}\mathbf{X}^v - \lambda_2 \mathbf{E}^{v,\mathrm{T}}(\mathbf{L}_s^v + \mathbf{L}_c)\mathbf{X}^v - \lambda_3 \mathbf{E}^{v,\mathrm{T}}(\widetilde{\boldsymbol{\Lambda}}_s^v + \widetilde{\boldsymbol{\Lambda}}_c)\mathbf{X}^v\right) \tag{12}$$

2) Update $\mathbf{H}_s^v$ with $\mathbf{U}^v, \mathbf{B}_s^v, \mathbf{H}_c, \mathbf{B}_c^v$ fixed
When fixing the other variables except $\mathbf{H}_s^v$, the minimization problem is defined below:

$$\min_{\mathbf{H}_s^v} \sum_{v=1}^{V}\left\|\widetilde{\mathbf{X}}^v - \mathbf{H}_s^{v,\mathrm{T}}\mathbf{B}_s^v - \mathbf{H}_c^{\mathrm{T}}\mathbf{B}_c^v\right\|_F + \lambda_1 \sum_{v=1}^{V}\left\|\mathbf{H}_s^{v,\mathrm{T}}\mathbf{H}_c\right\|_F \tag{13}$$

By taking the derivative of (13) with respect to $\mathbf{H}_s^v$ and setting it to zero, the rules for updating $\mathbf{H}_s^v$ is shown below:

$$\mathbf{H}_s^v = \left(\mathbf{B}_s^v\mathbf{B}_s^{v,\mathrm{T}} + \lambda_1 \mathbf{H}_c\mathbf{H}_c^{\mathrm{T}}\right)^{-1}\left(\mathbf{B}_s^v\widetilde{\mathbf{X}}^{v,\mathrm{T}} - \mathbf{B}_s^v\mathbf{B}_c^{v,\mathrm{T}}\mathbf{H}_c\right) \tag{14}$$

3) Update $\mathbf{B}_s^v$ with $\mathbf{U}^v, \mathbf{H}_s^v, \mathbf{H}_c, \mathbf{B}_c^v$ fixed
When fixing the other variables except $\mathbf{B}_s^v$, the minimization problem is defined below:

$$\min_{\mathbf{B}_s^v} \sum_{v=1}^{V}\left\|\widetilde{\mathbf{X}}^v - \mathbf{H}_s^{v,\mathrm{T}}\mathbf{B}_s^v - \mathbf{H}_c^{\mathrm{T}}\mathbf{B}_c^v\right\|_F \tag{15}$$

By taking the derivative of (15) with respect to $\mathbf{B}_s^v$ and setting it to zero, the rules for updating $\mathbf{B}_s^v$ is given below:

$$\mathbf{B}_s^v = \left(\mathbf{H}_s^v\mathbf{H}_s^{v,\mathrm{T}}\right)^{-1}\left(\mathbf{H}_s^v\widetilde{\mathbf{X}}^v - \mathbf{H}_s^v\mathbf{H}_c^{\mathrm{T}}\mathbf{B}_c^v\right) \tag{16}$$

4) Update $\mathbf{H}_c$ with $\mathbf{U}^v, \mathbf{H}_s^v, \mathbf{B}_s^v, \mathbf{B}_c^v$ fixed
When fixing the other variables except $\mathbf{H}_c$, the minimization problem is defined below:

$$\min_{\mathbf{H}_c} \sum_{v=1}^{V}\left\|\widetilde{\mathbf{X}}^v - \mathbf{H}_s^{v,\mathrm{T}}\mathbf{B}_s^v - \mathbf{H}_c^{\mathrm{T}}\mathbf{B}_c^v\right\|_F + \lambda_1 \sum_{v=1}^{V}\left\|\mathbf{H}_s^{v,\mathrm{T}}\mathbf{H}_c\right\|_F \tag{17}$$

By taking the derivative of (17) with respect to $\mathbf{H}_c$ and setting it to zero, the rules for updating $\mathbf{H}_c$ is given below:

$$\mathbf{H}_c = \left(\sum_{v=1}^{V}\mathbf{B}_c^v\mathbf{B}_c^{v,\mathrm{T}} + \lambda_1 \sum_{v=1}^{V}\mathbf{H}_s^v\mathbf{H}_s^{v,\mathrm{T}}\right)^{-1}\left(\sum_{v=1}^{V}\left(\mathbf{B}_c^v\widetilde{\mathbf{X}}^{v,\mathrm{T}} - \mathbf{B}_c^v\mathbf{B}_s^{v,\mathrm{T}}\mathbf{H}_s^v\right)\right) \tag{18}$$

5) Update $\mathbf{B}_c^v$ with $\mathbf{U}^v, \mathbf{H}_s^v, \mathbf{B}_s^v, \mathbf{H}_c$ fixed
When fixing the other variables except $\mathbf{B}_c^v$, the minimization problem is defined below:

$$\min_{\mathbf{B}_c} \sum_{v=1}^{V}\left\|\widetilde{\mathbf{X}}^v - \mathbf{H}_s^{v,\mathrm{T}}\mathbf{B}_s^v - \mathbf{H}_c^{\mathrm{T}}\mathbf{B}_c^v\right\|_F \tag{19}$$

By taking the derivative of (19) with respect to $\mathbf{B}_c^v$ and setting it to zero, the rules for updating $\mathbf{B}_c^v$ is given below:

$$\mathbf{B}_c^v = (\mathbf{H}_c\mathbf{H}_c^{\mathrm{T}})^{-1}\left(\mathbf{H}_c \sum_{v=1}^{V}\widetilde{\mathbf{X}}^v - \mathbf{H}_c \sum_{v=1}^{V}\mathbf{H}_s^{v,\mathrm{T}}\mathbf{B}_s^v\right) \tag{20}$$

The locally optimal solution of (10) is achieved by iterative updating the above parameters by performing (12), (14), (16), (18) and (20).

In the testing process, we let $\mathbf{B}_s^{*,v}$ and $\mathbf{B}_c^{*,v}$ as the learned base matrix of $v$-th view from the training process, and only the error matrix $\mathbf{U}_{te}^v$, common representation $\mathbf{H}_{c,te}$ and specific representation $\mathbf{H}_{s,te}^v$ for test data need to be learned. Let $\{\mathbf{X}_{te}^v \in R^{N_{te} \times d^v}, v=1,2,\ldots,V\}$ as test data, $\mathbf{E}_{te}^v \in R^{N_{te} \times N_{te}}$ be the indicator matrix for the test data. In addition, $\mathbf{L}_{s,te}^v$, $\mathbf{L}_{c,te}$, $\mathbf{\Lambda}_{s,te}^v$, $\mathbf{\Lambda}_{c,te}$ are estimated from test data. Drawing on the previously mentioned variables, the objective function for test data is given below:

$$\min_{\mathbf{U}^v, \mathbf{H}_s^v, \mathbf{H}_c} \sum_{v=1}^V \left\| \widetilde{\mathbf{X}}_{te}^v - \mathbf{H}_{s,te}^{v,\mathrm{T}} \mathbf{B}_s^{*,v} - \mathbf{H}_{c,te}^{\mathrm{T}} \mathbf{B}_c^{*,v} \right\|_F + \lambda_1 \sum_{v=1}^V tr(\mathbf{H}_{s,te}^{v,\mathrm{T}} \mathbf{H}_{c,te}) + \lambda_2 \left(\sum_{v=1}^V tr(\widetilde{\mathbf{X}}_{te}^{v,\mathrm{T}} \mathbf{L}_{s,te}^v \widetilde{\mathbf{X}}_{te}^v)\right) + \sum_{v=1}^V tr\left(\widetilde{\mathbf{X}}_{te}^{v,\mathrm{T}} \mathbf{L}_{c,te} \widetilde{\mathbf{X}}_{te}^v\right)\right) + \lambda_3 \left(\sum_{v=1}^V tr(\widetilde{\mathbf{X}}_{te}^{v,\mathrm{T}} \mathbf{\Lambda}_{s,te}^{v,\mathrm{T}} \mathbf{\Lambda}_{s,te}^v \widetilde{\mathbf{X}}_{te}^v) + \sum_{v=1}^V tr(\widetilde{\mathbf{X}}_{te}^{v,\mathrm{T}} \mathbf{\Lambda}_{c,te}^{\mathrm{T}} \mathbf{\Lambda}_{c,te} \widetilde{\mathbf{X}}_{te}^v)\right) \quad (21)$$

where $\widetilde{\mathbf{X}}_{te}^v = \mathbf{X}_{te}^v + \mathbf{E}_{te}^v \mathbf{U}_{te}^v$. Similar to (12), (14) and (16), denoting $\widetilde{\mathbf{\Lambda}}_c = \mathbf{\Lambda}_c^{\mathrm{T}} \mathbf{\Lambda}_c$, $\widetilde{\mathbf{\Lambda}}_s^v = \mathbf{\Lambda}_s^{v,\mathrm{T}} \mathbf{\Lambda}_s^v$, the iterative rules for solving $\mathbf{U}_{te}^v$, $\mathbf{H}_{s,te}^v$ and $\mathbf{H}_{c,te}$ for test data are given by:

$$\mathbf{U}_{te}^v = \left(\mathbf{E}_{te}^{v,\mathrm{T}} \mathbf{E}_{te}^v + \lambda_2 \mathbf{E}_{te}^{v,\mathrm{T}} (\mathbf{L}_{s,te}^v + \mathbf{L}_{c,te}) \mathbf{E}_{te}^v + \lambda_3 \mathbf{E}_{te}^{v,\mathrm{T}} (\widetilde{\mathbf{\Lambda}}_{s,te}^v + \widetilde{\mathbf{\Lambda}}_{c,te}) \mathbf{E}_{te}^v\right)^{-1}$$
$$\left(\mathbf{E}_{te}^{v,\mathrm{T}} \mathbf{H}_{s,te}^{v,\mathrm{T}} \mathbf{B}_s^{*,v} + \mathbf{E}_{te}^{v,\mathrm{T}} \mathbf{H}_{c,te}^{\mathrm{T}} \mathbf{B}_c^{*,v} - \mathbf{E}_{te}^{v,\mathrm{T}} \mathbf{X}_{te}^v - \lambda_2 \mathbf{E}_{te}^{v,\mathrm{T}} (\mathbf{L}_{s,te}^v + \mathbf{L}_{c,te}) \mathbf{X}_{te}^v - \lambda_3 \mathbf{E}_{te}^{v,\mathrm{T}} (\widetilde{\mathbf{\Lambda}}_{s,te}^v + \widetilde{\mathbf{\Lambda}}_{c,te}) \mathbf{X}_{te}^v\right) \quad (22)$$

$$\mathbf{H}_{s,te}^v = \left(\mathbf{B}_s^{*,v} \mathbf{B}_s^{*,v,\mathrm{T}} + \lambda_1 \mathbf{H}_{c,te} \mathbf{H}_{c,te}^{\mathrm{T}}\right)^{-1} \left(\mathbf{B}_s^{*,v} \widetilde{\mathbf{X}}_{te}^{v,\mathrm{T}} - \mathbf{B}_s^{*,v} \mathbf{B}_c^{*,v,\mathrm{T}} \mathbf{H}_{c,te}\right) \quad (23)$$

$$\mathbf{H}_{c,te} = \left(\mathbf{B}_c^{*,v} \mathbf{B}_c^{*,v,\mathrm{T}} + \lambda_1 \sum_{v=1}^V \mathbf{H}_{s,te}^v \mathbf{H}_{s,te}^{v,\mathrm{T}}\right)^{-1} \left(\sum_{v=1}^V \left(\mathbf{B}_c^{*,v} \widetilde{\mathbf{X}}_{te}^{v,\mathrm{T}} - \mathbf{B}_c^{*,v} \mathbf{B}_s^{*,v,\mathrm{T}} \mathbf{H}_{s,te}^v\right)\right) \quad (24)$$

Hence, the training stage of Incomplete Multi-View Dual Representation Learning (IMV_DRL) is presented in detail in Algorithm 1, and Algorithm S-1 (test) in the *Supplementary Materials* section describes the testing process for IMV_DRL.

---

**Algorithm 1** IMV_DRL (train)

---

**Input**: incomplete multi-view data $\mathbf{X}^v$, $v=1,2\ldots V$. The regularization hyperparameters $\lambda_1, \lambda_2, \lambda_3$. Maximum iterations $T$.

---

1: Randomly initialize $\mathbf{E}^v, \mathbf{H}_s^v, \mathbf{B}_s^v, \mathbf{H}_c$ and $\mathbf{B}_c^v$, $v=1,2\ldots V$.
2: while $t<T$ do
3:   while $v<V$ do
4:     Update $\mathbf{U}^v$ according to (12).
5:     Update $\mathbf{H}_s^v$ according to (14).
6:     Update $\mathbf{B}_s^v$ according to (15).
7:     Update $\mathbf{B}_c^v$ according to (20).
8:   end while
9:   Update $\mathbf{H}_c$ according to (18).
10:  if (10) converged
11:    return.
12:  end if
13: end while

---

**Output**: the error matrix $\mathbf{U}^v$, common representation $\mathbf{H}_c$ and sepcific representation $\mathbf{H}_s^v$, v=1,2…V.

---

### 3.3 Incomplete Multi-view Fuzzy System with Dual Hidden Representation Cooperative Learning

Recent advancements have led to the proposal of several effective methods for modeling multi-view fuzzy systems. For instance, a representative multi-view fuzzy system is constructed in [18] and the objective function is as follows:

$$\min_{\mathbf{P}_g^v, \alpha^v} \sum_{v=1}^V \alpha^v \left\| \mathbf{X}_g^v \mathbf{P}_g^v - \mathbf{Y} \right\|_2 + \beta \sum_{v=1}^V \left\| \mathbf{X}_g^v \mathbf{P}_g^v - \frac{1}{V-1} \sum_{l=1, l \neq v}^V \mathbf{X}_g^l \mathbf{P}_g^l \right\|_2 + \gamma \sum_{v=1}^V \alpha^v \ln \alpha^v + \delta \sum_{v=1}^V \left\| \mathbf{P}_g^v \right\|_2$$
$$s.t.\ \alpha^v \geq 0, \sum_{v=1}^V \alpha^v = 1 \quad (25)$$

where $\mathbf{X}_g^v \in R^{N \times d_g^v}$ is fuzzy feature space, which is obtained by mapping $\mathbf{X}^v$ with fuzzy rules, $d_g^v$ is the number of fuzzy feature dimensions. $\mathbf{P}_g^v \in R^{d_g^v \times c}$ is the consequent parameters of the $v$-th view. $\mathbf{Y} = [\mathbf{y}_1; \mathbf{y}_2; \ldots; \mathbf{y}_N] \in R^{N \times c}$ is the label matrix, and $\mathbf{y}_i \in R^{1 \times c}$ is the label vector for the $i$-th sample. $\alpha^v$ is the view weight associated with view $v$. $\beta$, $\gamma$, and $\delta$ are the regularization parameters. In (25), the first term is used to learn the consequent parameters of each view, the second term is used to explore the consistent information among views, the third term is introduced to adjust the view weight, and the fourth term is the regularization term.

Although (25) has received great performance, this method still cannot address incomplete multi-view data directly. Besides, (25) only explores the information of visible data to model, but neglects the hidden information between views. Therefore, a new effective fuzzy system for incomplete multi-view data is needed.

In this subsection, a novel TSK fuzzy system with cooperative learning between dual hidden views and complete visible views (DRIMV_TSK) is proposed. Specifically, we denote the specific view as $\mathbf{Z}_s = [\mathbf{H}_s^1, \mathbf{H}_s^2, \ldots, \mathbf{H}_s^V] \in R^{N \times (V \times m_s)}$, the common view as $\mathbf{Z}_c = \mathbf{H}_c \in R^{N \times m_c}$, and let them as the ($V$+1)-th and the ($V$+2)-th view. Then the dual hidden views and imputed multi-view data are unified to construct the model. Based on the above analyses, the objective function of DRIMV_TSK is defined as below:

$$\min_{\mathbf{P}_g^v, \alpha^v} \sum_{v=1}^{V} \alpha^v \|\widetilde{\mathbf{X}}_g^v \mathbf{P}_g^v - \mathbf{Y}\|_2 + \alpha^{V+1} \|\mathbf{Z}_{c,g} \mathbf{P}_g^{V+1} - \mathbf{Y}\|_2 + \alpha^{V+2} \|\mathbf{Z}_{s,g} \mathbf{P}_g^{V+2} - \mathbf{Y}\|_2 + \beta \left( \sum_{v=1}^{V} \|\widetilde{\mathbf{X}}_g^v \mathbf{P}_g^v - \mathbf{\Lambda}\|_F + \|\mathbf{Z}_{c,g} \mathbf{P}_g^{V+1} - \mathbf{\Lambda}\|_F + \|\mathbf{Z}_{s,g} \mathbf{P}_g^{V+2} - \mathbf{\Lambda}\|_F \right) + \gamma \sum_{v=1}^{V+2} \alpha^v \ln \alpha^v + \delta \left( \sum_{v=1}^{V+2} \|\mathbf{P}_g^v\|_2 + \|\mathbf{F}\|_2 \right)$$
$$s.t. \ \alpha^v \geq 0, \sum_{v=1}^{V} \alpha^v = 1 \tag{26}$$

where $\widetilde{\mathbf{X}}_g^v \in R^{N \times d_g^v}$ is the mapping of the imputed data $\widetilde{\mathbf{X}}^v$ in the fuzzy feature space by fuzzy rules. $\mathbf{Z}_{s,g} \in R^{N \times d_g^{V+2}}$ and $\mathbf{Z}_{c,g} \in R^{N \times d_g^{V+1}}$ are the mapping of the specific view $\mathbf{Z}_s$ and common view $\mathbf{Z}_c$ in the fuzzy feature space by fuzzy rules. (26) is described in detail as below:

1) The first three terms $\sum_{v=1}^{V} \alpha^v \|\widetilde{\mathbf{X}}_g^v \mathbf{P}_g^v - \mathbf{Y}\|_2 + \alpha^{V+1} \|\mathbf{Z}_{c,g} \mathbf{P}_g^{V+1} - \mathbf{Y}\|_2 + \alpha^{V+2} \|\mathbf{Z}_{s,g} \mathbf{P}_g^{V+2} - \mathbf{Y}\|_2$ are the training error of all views.

2) The fourth to sixth term $\sum_{v=1}^{V} \|\widetilde{\mathbf{X}}_g^v \mathbf{P}_g^v - \mathbf{\Lambda}\|_F + \|\mathbf{Z}_{c,g} \mathbf{P}_g^{V+1} - \mathbf{\Lambda}\|_F + \|\mathbf{Z}_{s,g} \mathbf{P}_g^{V+2} - \mathbf{\Lambda}\|_F$ are the cooperation terms, which are used to align the output of all views, and $\mathbf{\Lambda} = \sum_{l,l \neq v}^{V+2} \widetilde{\mathbf{X}}_g^l \mathbf{P}_g^l$.

3) The seventh term $\sum_{v=1}^{V+2} \alpha^v \ln \alpha^v$ is the negative Shannon entropy. Following [18, 37], the importance of variables can be adaptively learned by introducing the negative entropy term into the objective function. Based on this term, the optimal view weight can be obtained.

4) $\beta$, $\gamma$, and $\delta$ are the regularization hyperparameters that are introduced to balance the effects of the corresponding terms.

Similarly, the iterative optimization method is used to solve (26). Denoting $\mathbf{Z}_{c,g}$ and $\mathbf{Z}_{s,g}$ as the ($V$+1)-th and the ($V$+2)-th view. The updated rules of $\mathbf{P}_g^v$ and $\alpha^v$ are given as follows:

$$\mathbf{P}_g^v = \left( (\alpha^v + \beta) \widetilde{\mathbf{X}}_g^{v,T} \widetilde{\mathbf{X}}_g^v + \delta \mathbf{I} \right)^{-1} \left( \widetilde{\mathbf{X}}_g^{v,T} \mathbf{Y} + \beta \widetilde{\mathbf{X}}_g^{v,T} \mathbf{\Lambda} \right)$$
$$v = 1, 2, \ldots, V+2 \tag{27}$$

$$\alpha^v = \frac{\exp\left(-\|\widetilde{\mathbf{X}}_g^v \mathbf{P}_g^v - \mathbf{Y}\|_2 / \gamma\right)}{\sum_{l=1}^{V+2} \exp\left(-\|\widetilde{\mathbf{X}}_g^l \mathbf{P}_g^l - \mathbf{Y}\|_2 / \gamma\right)} \quad v = 1, 2, \ldots, V+2 \tag{28}$$

where $\mathbf{I} \in R^{d_g^v \times d_g^v}$. The locally optimal solution of (26) is found by iterative implementing the above two updating steps.

The details of the propsoed method are shown in Algorithm 2. By performing Algorithm 1, the imputed visible data and dual hidden representation of test dataset $\{\mathbf{X}_{te}^v \in R^{N \times d^v}, v = 1, 2, \ldots, V\}$ can be obtained. After it, mapping the the imputed visible data and dual hidden representation into fuzzy feature space by using (3b)-(3d). Finally, the calculation results are output as below:

$$\mathbf{Y}_{output} = \sum_{v=1}^{V} \alpha^v \widetilde{\mathbf{X}}_{te,g}^v \mathbf{P}_g^v + \alpha^{V+1} \mathbf{Z}_{c,te,g} \mathbf{P}_g^{V+1} + \alpha^{V+2} \mathbf{Z}_{s,te,g} \mathbf{P}_g^{V+2} \tag{29}$$

---

**Algorithm 2** DRIMV_TSK

**Input**: incomplete multi-view data $\mathbf{X}^v$, $v$=1, 2, …, $V$, number of fuzzy rule $K$, the maximum number of iterations $T$, regularization hyperparameters $\beta$, $\gamma$ and $\delta$.
**Output**: $\mathbf{P}_g^v$ and $\alpha^v$, $v$=1, 2, …, $V$+1.

1: Learning the imputed multi-view data $\widetilde{\mathbf{X}}^v$, the common hidden information $\mathbf{H}_c$ and specific hidden information $\mathbf{H}_s^v$ using Algorithm1.
2: Evaluating the antecedent parameters for multi-view data by using Var-Part clustering algorithm.
3: mapping the data into fuzzy feature space, and constructing a new multi-view dataset $D^{V+2} = \{\{\widetilde{\mathbf{X}}_g^v, \mathbf{Z}_{c,g}, \mathbf{Z}_{s,g}\}, \mathbf{Y}\}$ based on (3b) - (3d).
4: Randomly initialize $\mathbf{P}_g^v$, and $\alpha^v = 1/(V+2)$.
5: while $t < T$ do
6:   while $v < V+2$ do
7:     Updating $\mathbf{P}_g^v$ by using (27).
8:     Updating $\alpha^v$ by using (28).
9:   end while
10: if (26) converged
11:     Break
12: end if
13: end while

---

### 3.4 Complexity Analysis

In this subsection, the computational complexity analyses of the proposed DRIMV_TSK are discussed in detail. In Algorithm 2, Step 1 is the IMV_DRL algorithm, and it includes five update steps. Denoting $N$ as the number of instances, $M$ as the max hidden representation dimension, $D$ as the max origin view dimension, $V$ as the number of views, and $T$ as the number of iterations. In IMV_DRL, the complexity of updating $\mathbf{U}^v$ is $O((N^2 + NM + MD + N)NVT)$, the complexity of updating $\mathbf{H}_s^v$ is $O((MN +$ $MD + DN)MVT\big)$, the complexity of updating $\mathbf{B}_s^v$ and $\mathbf{B}_c^v$ are $O\big((MN + MD + DN)MVT\big)$, and the complexity of updating $\mathbf{H}_c$ is $O\big((MN + MD + DN)MT\big)$. Therefore, the complexity of step 1 is $O(N^3VT)$. The complexity of step 2 and 3 are $O(2NDK)$ and $O\big(NK(D+1)\big)$, respectively. where $K$ is the number of rules. In addition, the complexity of steps 7 and 8 are $O\big((M_gN + NC + M_gC)M_g(V+2)T\big)$ and $O\big((NM_g + CM_g + CN)C(V+2)T\big)$, respectively. $M_g$ is the max feature dimension in fuzzy space, and $C$ is the number of classes. Finally, the overall computational cost of Algorithm 2 is $O(N^3VT)$.

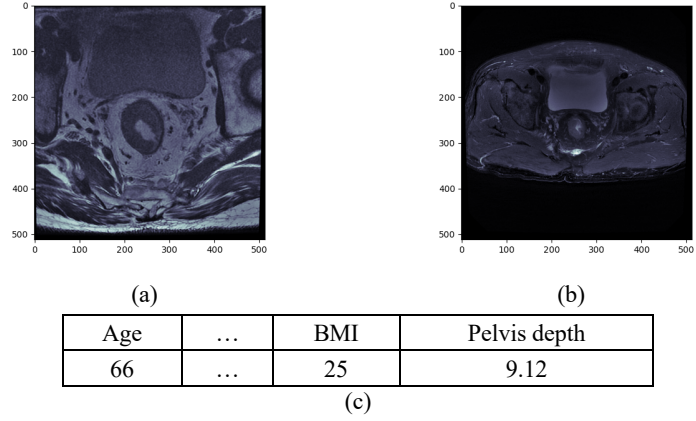

| Age | … | BMI | Pelvis depth |
|---|---|---|---|
| 66 | … | 25 | 9.12 |

(c)

**Fig. 2**. A patient in MVCR dataset with three views. (a) is the high-resolution MRI view, (b) is the pressed-fat MRI view, and (c) is the clinical view.

## 4  Experimental Studies

### 4.1  Data Construction and Processing

Up to now, there are no published datasets specifically for assessing surgical difficulty in rectal cancer. To address this gap, a new multi-view rectal cancer dataset, referred to as MVCR, has been developed using patient data before October 23, 2023 from the Affiliated Hospital of Jiangnan University. Specifically, this study was a single-center study with patients retrospectively enrolled from Affiliated Hospital of Jiangnan University, and it was approved by the Ethics Committee of the participating hospital, with the requirement for informed consent waived. MVCR consists of 151 patients, and three views (high-resolution and pressed-fat MRI image view, traditional clinical data view). Fig. 2 shows one patient's data with three views. There are two classes in the MVCR, i.e., easy and difficult, which presents the surgical difficulty as low or high, and they are given by the doctors.

Then, based on the pyradiomics toolkit, the radiomics feature of two types of MRI images is extracted. Specifically, we extract their first-order features, shape features, and second-order features, when the wavelet variation has been conducted on both two types of MRI images. Then, we remove the features with zero variance and obtain the final radiomics feature. Finally, combining the clinical data view, we obtained the MVCR dataset for evaluation.

### 4.2  Methods for Comparison

To effectively evaluate the effectiveness of the proposed DRIMV_TSK, five incomplete multi-view learning methods and two traditional multi-view classification methods are compared with DRIMV_TSK. Specifically, we conduct three imputation algorithms on two traditional multi-view learning methods to enable them to address incomplete multi-view data. The methods are briefly described as follows.

*Imputation algorithms*:

Mean imputation method: The imputation of missing views was done by using the average values of the complete views.

KNN imputation method: the missing views were imputed with the average values of $k$ most similar feature values, which is found by using the KNN algorithm [26].

SVT imputation method: the missing views were imputed by using the low-rank approximation algorithm [25].

*Traditional multi-view learning methods*:

TwoV-TSK : this algorithm is based on TSK and introduces cooperative learning with maximum marginal learning criterion for multi-view learning [41].

AMVMED: this algorithm is based on maximum entropy discrimination to explore the consistent information between views for modeling [9].

*Incomplete multi-view learning methods*:

IMG: this algorithm integrated NMF and manifold learning to extract a common representation, and then used the TSK fuzzy system to model it [13].

IMC_GRMF: this algorithm integrated NMF and graph regularization to extract a common representation, and then the TSK fuzzy system is conducted on common representation for modeling [14].

DAIMC: this algorithm integrated semi-NMF to extract common representation, and then the TSK fuzzy system is conducted on common representation for modeling [29].

IMSF: this algorithm constructed a classification model by partitioning the available data into multiple complete blocks of data for modeling [33].

IMV_TSK: this algorithm integrated common representation learning and missing view imputation as a unified optimization framework. After that, this algorithm combined these two types of data to model [37].

### 4.3 Experimental Settings

To be fair, regularization parameters of all methods were set in $\{2^{-5}, 2^{-4}, \ldots, 2^{4}, 2^{5}\}$. The number of fuzzy rules was set in [2, 4, 5, 8, 10]. The instances in the MVCR dataset were randomly removed from 10% to 70% with a step size of 20%. The above procedures were performed ten times and the mean accuracy and variances were recorded for comparison to minimize potential bias effects. To effectively evaluate the performance of algorithms, three metrics, ACC, AUC, and F1, are used in the experiments. It is important be noted that AMVMED, TwoV-TSKFS, and IMC_GRMF_TSK are limited to handing two view datasets. Therefore, these methods were performed on any two view data, and their optimal accuracy were recorded for comparison.

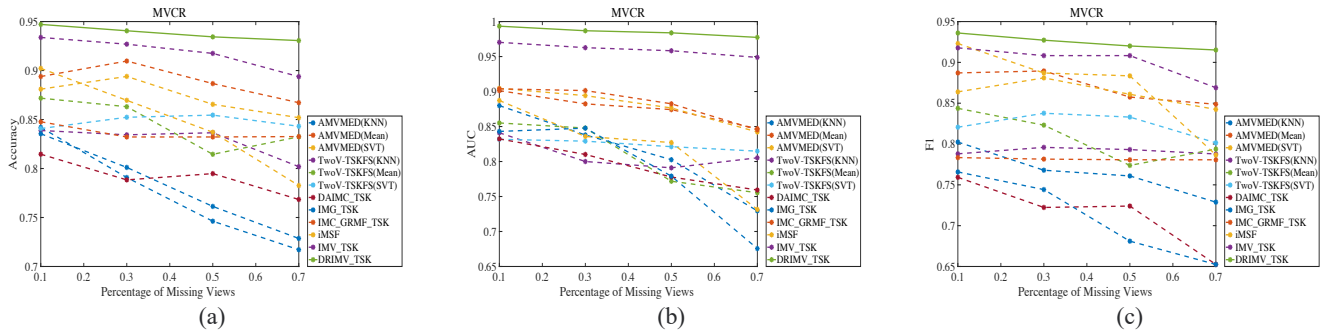

**Fig. 3.** Experimental results of all algorithms on MVCR dataset.

**Table 1.** Performance of all algorithms with 10% rate of missing views.

| Algorithm | ACC | AUC | F1 |
| --- | --- | --- | --- |
| AMVMED(Mean) | 0.8420±0.0128 | 0.8799±0.0143 | 0.8024±0.0068 |
| AMVMED(KNN) | 0.8939±0.0134 | 0.9013±0.0420 | 0.8871±0.0172 |
| AMVMED(SVT) | 0.8810±0.0067 | 0.9045±0.0308 | 0.8639±0.0091 |
| TwoV-TSKFS(Mean) | 0.8388±0.0300 | 0.8411±0.0339 | 0.7880±0.0434 |
| TwoV-TSKFS(KNN) | 0.8718±0.0098 | 0.8551±0.0243 | 0.8437±0.0130 |
| TwoV-TSKFS(SVT) | 0.8408±0.0268 | 0.8317±0.0194 | 0.8206±0.0251 |
| DAIMC_TSK | 0.8146±0.0555 | 0.8323±0.0912 | 0.7592±0.0704 |
| IMG_TSK | 0.8355±0.0323 | 0.8430±0.0912 | 0.7658±0.0279 |
| IMC_GRMF_TSK | 0.8477±0.0197 | 0.9038±0.0023 | 0.7834±0.0178 |
| IMSF | 0.9022±0.0153 | 0.8872±0.0221 | 0.9231±0.0231 |
| IMV_TSK | 0.9338±0.0333 | 0.9703±0.0015 | 0.9178±0.0034 |
| **DRIMV_TSK** | **0.9471±0.0009** | **0.9934±0.0007** | **0.9360±0.0019** |

### 4.4 Experimental Results

The performance of all methods is shown in Tables 1-4 and Fig. 3. By thoroughly analyzing the data presented in these results, the following conclusions can be reached.

1) First, DRIMV_TSK achieves better performance compared with most methods in most cases. This indicates that exploring and integrating common and specific information with imputed multi-view data for modeling is effective. Besides, since the proposed DRIMV_TSK is built upon the interpretable fuzzy system, it still has advantages in interpretability.

2) When the rate of missing views is relatively small, the imputation based methods can obtain good performance, but the accuracy of these methods degrades with the increase of missing views. Therefore, some new mechanism is needed, such as combining missing view imputation and representation learning into a unified optimization process.

3) Similarly, the group based method, IMSF, achieves great classification accuracy when there are relatively few missing views. Since the available information in each group of data becomes scarce as the missing rate becomes larger, the classification accuracy of the IMSF grate degrades.

4) Common representation based methods cannot achieve better performance in most cases. This is because these methods only use common information for modeling and ignore other information, such as specific information in each view, and it fails to maintain strong classification performance when the rate of missing views increase.

5) IMV_TSK achieves a better performance, which indicates that combining data imputation and hidden representation learning for modeling is useful. However, IMV_TSK only extracts the common information and only considers the one-order similar learning, which leads to the performance of IMV_TSK inferiors to DRIMV_TSK in most cases.

**Table 2.** Performance of all algorithms with 30% rate of missing views.

| Algorithm | ACC | AUC | F1 |
|---|---|---|---|
| AMVMED(Mean) | 0.7908±0.0035 | 0.8380±0.0271 | 0.7679±0.0413 |
| AMVMED(KNN) | 0.9097±0.0038 | 0.8824±0.0087 | 0.8895±0.0124 |
| AMVMED(SVT) | 0.8940±0.0135 | 0.8942±0.0264 | 0.8810±0.0049 |
| TwoV-TSKFS(Mean) | 0.8344±0.0064 | 0.8000±0.0479 | 0.7959±0.0291 |
| TwoV-TSKFS(KNN) | 0.8631±0.0102 | 0.8478±0.0215 | 0.8231±0.0215 |
| TwoV-TSKFS(SVT) | 0.8523±0.0139 | 0.8291±0.0473 | 0.8377±0.0249 |
| DAIMC_TSK | 0.7884±0.0588 | 0.8101±0.0104 | 0.7223±0.0741 |
| IMG_TSK | 0.8011±0.0481 | 0.8477±0.0490 | 0.7442±0.0620 |
| IMC_GRMF_TSK | 0.8322±0.0039 | 0.9014±0.0036 | 0.7815±0.0069 |
| IMSF | 0.8696±0.0923 | 0.8358±0.0116 | 0.8870±0.0313 |
| IMV_TSK | 0.9269±0.0420 | 0.9627±0.0076 | 0.9083±0.0040 |
| DRIMV_TSK | **0.9406±0.0002** | **0.9869±0.0067** | **0.9272±0.0004** |

**Table 3.** Performance of all algorithms with 50% rate of missing views.

| Algorithm | ACC | AUC | F1 |
|---|---|---|---|
| AMVMED(Mean) | 0.7462±0.0150 | 0.8026±0.0111 | 0.7609±0.0205 |
| AMVMED(KNN) | 0.8867±0.0246 | 0.8742±0.0194 | 0.8575±0.0076 |
| AMVMED(SVT) | 0.8655±0.0140 | 0.8765±0.0283 | 0.8609±0.0139 |
| TwoV-TSKFS(Mean) | 0.8365±0.0155 | 0.7910±0.0112 | 0.7932±0.0215 |
| TwoV-TSKFS(KNN) | 0.8145±0.0040 | 0.7718±0.0326 | 0.7738±0.0045 |
| TwoV-TSKFS(SVT) | 0.8545±0.0292 | 0.8212±0.0326 | 0.8330±0.0371 |
| DAIMC_TSK | 0.7948±0.0424 | 0.7778±0.0104 | 0.7240±0.0631 |
| IMG_TSK | 0.7613±0.0450 | 0.7791±0.0747 | 0.6810±0.0772 |
| IMC_GRMF_TSK | 0.8322±0.0041 | 0.8825±0.0082 | 0.7807±0.0105 |
| IMSF | 0.8370±0.0176 | 0.8271±0.0937 | 0.8835±0.0692 |
| IMV_TSK | 0.9175±0.0420 | 0.9583±0.0041 | 0.9083±0.0140 |
| DRIMV_TSK | **0.9344±0.0036** | **0.9839±0.0016** | **0.9201±0.0012** |

**Table 4.** Performance of all algorithms with 70% rate of missing views.

| Algorithm | ACC | AUC | F1 |
|---|---|---|---|
| AMVMED(Mean) | 0.7171±0.0211 | 0.7295±0.3050 | 0.7288±0.0372 |
| AMVMED(KNN) | 0.8672±0.0070 | 0.8470±0.0346 | 0.8487±0.0411 |
| AMVMED(SVT) | 0.8519±0.0314 | 0.8422±0.0226 | 0.8424±0.0224 |
| TwoV-TSKFS(Mean) | 0.8019±0.0176 | 0.8053±0.0366 | 0.7878±0.0434 |
| TwoV-TSKFS(KNN) | 0.8324±0.0235 | 0.7554±0.0337 | 0.7938±0.0284 |
| TwoV-TSKFS(SVT) | 0.8431±0.0408 | 0.8148±0.0564 | 0.8015±0.0582 |
| DAIMC_TSK | 0.7684±0.0392 | 0.7593±0.0109 | 0.6525±0.0137 |
| IMG_TSK | 0.7286±0.0258 | 0.6756±0.0946 | 0.6525±0.0137 |
| IMC_GRMF_TSK | 0.8326±0.0079 | 0.8453±0.0228 | 0.7807±0.0105 |
| IMSF | 0.7826±0.0307 | 0.7315±0.0468 | 0.7867±0.0874 |
| IMV_TSK | 0.8938±0.0799 | 0.9488±0.0081 | 0.8690±0.0068 |
| DRIMV_TSK | **0.9306±0.0047** | **0.9774±0.0043** | **0.9152±0.0048** |

Table 5. classification performance of DRIMV_TSK1, DRIMV_TSK2, DRIMV_TSK3 and DRIMV_TSK

| Algorithm | ACC | AUC | F1 |
|---|---|---|---|
| DRIMV_TSK1 | 0.9273±0.0580 | 0.9864±0.0134 | 0.9104±0.0067 |
| DRIMV_TSK2 | 0.9340±0.0035 | **0.9885±0.0057** | 0.9196±0.0043 |
| DRIMV_TSK3 | 0.9140±0.0556 | 0.9828±0.0048 | 0.8922±0.0119 |
| DRIMV_TSK | **0.9406±0.0002** | 0.9869±0.0067 | **0.9272±0.0004** |

## 4.5 Ablation Studies

To better evaluate the effectiveness of dual representation learning and cooperative learning, we perform experiments using several variants of the DRIMV_TSK model. Specifically, we define DRIMV_TSK without a common representation as DRIMV_TSK1, DRIMV_TSK without a specific representation as DRIMV_TSK2, and DRIMV_TSK without cooperative learning as DRIMV_TSK3. The experimental results of these methods are presented in Table 5, under conditions with a missing view rate of 50%. From Table 5, the following conclusions can be made. First, all three variants—DRIMV_TSK1, DRIMV_TSK2, and DRIMV_TSK3—demonstrate inferior performance compared to DRIMV_TSK. This finding indicates that both dual hidden representation and cooperative learning contribute significantly to the effectiveness of rectal cancer surgical evaluation. Second, DRIMV_TSK2, which lacks specific representation, shows notably poorer performance compared to the other methods. This suggests that introducing specific information is crucial for accurate modeling and improves the model's ability to evaluate surgical difficulty effectively. Third, the performance of DRIMV_TSK1 and DRIMV_TSK3 is similar when evaluated using ACC and AUC metrics. However, the F1 metric reveals that cooperative learning plays a more critical role than common representation in the modeling process. This indicates that cooperative learning has a greater impact on the model's overall performance compared to the shared representation.

Table 6. The Statistical value p and ranking of all algorithms on AUC metric

| Algorithms | Ranking | Statistical value p | Null Hypothesis |
|---|---|---|---|
| DRIMV_TSK | 1 | | |
| IMV_TSK | 2 | | |
| IMC_GRMF_TSK | 3.5 | | |
| AMVMED(SVT) | 4 | | |
| AMVMED(Mean) | 4.5 | | |
| IMSF | 7.75 | 0.00016 | Reject |
| AMVMED(KNN) | 8.5 | | |
| TwoV-TSKFS(Mean) | 8.75 | | |
| TwoV-TSKFS(SVT) | 8.75 | | |
| TwoV-TSKFS(KNN) | 9.5 | | |
| IMG_TSK | 9.5 | | |
| DAIMC_TSK | 10.25 | | |

Table 7. Post-holm results on AUC metric (reject hypothesis if statistical value $p < 0.0125$)

| $i$ | Algorithm | z | $p$ | Holm | Null Hypothesis |
|---|---|---|---|---|---|
| 11 | DAIMC_TSK / DRIMV_TSK | 3.628149 | 0.000285 | 0.004545 | Reject |
| 10 | IMG_TSK / DRIMV_TSK | 3.333974 | 0.000856 | 0.005 | Reject |
| 9 | TwoV-TSKFS(KNN) / DRIMV_TSK | 3.333974 | 0.000856 | 0.005556 | Reject |
| 8 | TwoV-TSKFS(SVT) / DRIMV_TSK | 3.039800 | 0.002367 | 0.00625 | Reject |
| 7 | TwoV-TSKFS(Mean) / DRIMV_TSK | 3.039800 | 0.002367 | 0.007143 | Reject |
| 6 | AMVMED(KNN) / DRIMV_TSK | 2.941742 | 0.003264 | 0.008333 | Reject |
| 5 | IMSF / DRIMV_TSK | 2.647568 | 0.008107 | 0.01 | Reject |
| 4 | AMVMED(Mean) / DRIMV_TSK | 1.372813 | 0.169811 | 0.0125 | Not Reject |
| 3 | AMVMED(SVT) / DRIMV_TSK | 1.176697 | 0.239317 | 0.016667 | Not Reject |
| 2 | IMC_GRMF_TSK / DRIMV_TSK | 0.980581 | 0.326800 | 0.025 | Not Reject |
| 1 | IMV_TSK / DRIMV_TSK | 0.392232 | 0.694887 | 0.05 | Not Reject |

### 4.6 Statistical Analysis

To further verify whether there is an essential difference between DRIMV_TSK and other incomplete multi-view classification methods. In this subsection, the Friedman and post-hoc Holm tests are conducted based on the results of AUC, ACC and F1 metrics in all rates of missing views.

At first, the Friedman test [42] is conducted to assess whether all algorithms have significant differences in terms of classification performance. We define the null hypothesis as that all methods have the same classification performance, and the null hypothesis is rejected when the statistical value $p<0.05$. The Friedman test results with AUC metric is shown in Table 6 and other results are given in S1-S2 of the *Supplementary Materials* section. It is evident from the results that all the statistical value $p$ is far less than 0.05, which indicates that all methods are remarkably differences. In addition, the proposed DRIMV_TSK ranks first, which indicates that DRIMV_TSK outperforms other algorithms.

Second, the post-hoc Holm test is further conducted to evaluate the difference between DRIMV_TSK and other methods. The Holm test results based on AUC metrics are given in Tables 7 and other results are given in S3-S4 of the *Supplementary Materials* section. When the statistical value $p<$Holm value, it indicates that there is a significantly difference between DRIMV_TSK and the corresponding algorithm. It can be seen from Tables 7 and S3-S4 that DRIMV_TSK is significant different from most of the methods. Despite the fact that DRIMV_TSK is not significantly different from a few methods, the classification results in Tables 1-4 and Fig. 3 also suggest that DRIMV_TSK is superior to these four algorithms in some way.

**Table 8.** Decision process of DRIMV_TSK in clinical view

---

**The first rule**:
**IF**: the age is *Little Large* and,
    the BMI is *Medium* and,
    …, and,
    the Pd is *Medium*.
**Then**: the 1th output is $0.4439 - 0.1001 x_{\text{age}} + 0.1646 x_{BMI} + \ldots - 0.9092 x_{\text{Pd}}$ and
    the 2th output is $-0.3302 + 0.1105 x_{\text{age}} - 0.08 x_{BMI} + \ldots + 0.9812 x_{\text{Pd}}$.

**The second rule**:
**IF**: the age is *Large* and,
    the BMI is *Large* and,
    …, and,
    the Pd is *Large*.
**Then**: the 1th output is $0.5118 - 0.1561 x_{\text{age}} + 0.4239 x_{BMI} + \ldots + 0.8013 x_{\text{Pd}}$ and
    the 2th output is $-0.4824 + 0.1042 x_{\text{age}} - 0.4152 x_{BMI} + \ldots - 0.7756 x_{\text{Pd}}$.

**The third rule**:
**IF**: the age is *Small* and,
    the BMI is *Small* and,
    …, and,
    the Pd is *Small*.
**Then**: the 1th output is $0.5137 - 0.1495 x_{\text{age}} + 0.2570 x_{BMI} + \ldots + 0.5175 x_{\text{Pd}}$ and
    the 2th output is $-0.4376 + 0.2096 x_{\text{age}} - 0.2193 x_{BMI} + \ldots - 0.4789 x_{\text{Pd}}$.

**The fourth rule**:
**IF**: the age is *Medium* and,
    the BMI is *Little Large* and,
    …, and,
    the Pd is *Little Large*.
**Then**: the 1th output is $0.3794 - 0.1463 x_{\text{age}} + 0.1692 x_{BMI} + \ldots - 1.0464 x_{\text{Pd}}$ and
    the 2th output is $-0.2865 + 0.1645 x_{\text{age}} - 0.0920 x_{BMI} + \ldots + 1.0930 x_{\text{Pd}}$.

---

### 4.7 Interpretability Analysis

To showcase the interpretability of DRIMV_TSK in evaluating surgical difficulty for rectal cancer, the model constructed with the rate of missing views at 50% is used as an example in this subsection. In the experiment, DRIMV_TSK extracts four fuzzy rules in clinical view data, which means that each feature dimension has four clustering centers. Similar to [37] and ordering the clustering center value, each fuzzy set can be transformed into four linguistic interpretations, i.e., *Small*, *Medium*, *Little Large*, and *Large*, respectively. The linguistic interpretation of fuzzy sets given above is one possible method of interpretation, and it can be defined as different linguistic interpretations in different application scenarios.

In this experiment, we choose the clinical view in the MVCR dataset to present the interpretability of DRIMV_TSK. Following

[37], the Gaussian function is chosen as the membership function. We use three dimensions of features to illustrate, i.e., Age, BMI, and Pelvis depth (Pd). Then, details of membership functions for each fuzzy set and possible linguistic interpretations are given in Fig. S1 of the *Supplementary Materials* section. As shown in the first row in Fig. S1, since the first clustering center of the Age feature is 0.7332 and it ranks second among the four centers (i.e., 0.7332, 0.7780, 0.2635 and 0.6699), the first fuzzy set can be transformed into *Little Large*. Similarly, the other feature dimensions can be transformed into the corresponding linguistic interpretations. Finally, the decision process of DRIMV_TSK can be explained by fuzzy rules, and the details of the decision process are shown in Table 8.

Fig. 4 further explains the decision process of DRIMV_TSK with a detailed example. It can be seen in Fig. 4 that rule 3 has a decisive effect in the assessment of rectal cancer surgical difficulty.

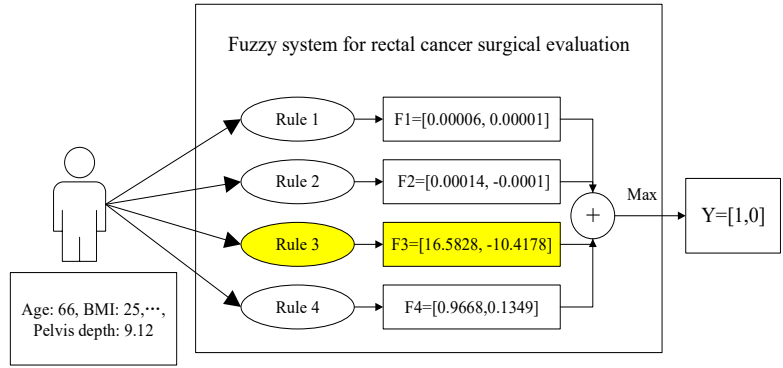

**Fig 4**. The decision-making process of the proposed DRIMV_TSK in a clinical context involves a combination operation, denoted by the symbol '+', and a 'Max' operation, which sets the largest element in *y* to 1 while setting all other elements to 0.

## 5  Conclusion

A multi-view rectal cancer dataset for surgical difficulty assessment is first constructed in this paper, which includes 150 patients. High-resolution and pressed-fat MRI images, and conventional clinical data are extracted for each patient as three views of the experiments. Then a new incomplete multi-view surgical difficulty assessment model is proposed based on this dataset. The proposed method includes two main components. The first is the incomplete multi-view dual representation learning method, which extracts common and specific information simultaneously and integrates missing view imputation and representation learning into one optimization process. In this framework, these two processes can negotiate with each other to learn the optimal dual hidden representation and imputed multi-view data. On the basis of the interpretable TSK fuzzy system, an incomplete multi-view model with dual representation cooperative learning is constructed in the second part. The proposed method balances performance and interpretability over existing methods. Extensive experimental results have demonstrated that DRIMV_TSK outperforms most existing incomplete multi-view models for rectal cancer surgical difficulty assessment.

Despite the promising results obtained by the proposed DRIMV_TSK, it still exists the following three issues. First, in the representation learning part, the proposed DRIMV_TSK can only explore the common and specific information as a linear relationship, and the performance of surgical difficulty assessment can be further improved if non-linear relationships can be explored by introducing deep learning. Second, While the proposed method is capable of handling cases with missing instances effectively, it falls short when dealing with scenarios involving missing features, which reduces its overall versatility in dealing with incomplete data. Third, it is a common problem in medical applications that there are few labeled data for modeling. The above three issues will be further addressed in further work. Fourth, how to introduce more robust learning mechanisms to further improve surgical assessment accuracy while ensuring interpretability remains an open question. Finally, how to collect more patient data and build a larger rectal cancer dataset is key to further advancing reliable surgical difficulty assessment.

**Acknowledgments**

This work was supported in part by National Key R&D Program of China under Grant (2022YFE0112400), the NSFC under Grant 62176105, the Six Talent Peaks Project in Jiangsu Province under Grant XYDXX-056. (Corresponding author: Zhaohong Deng and Shudong Hu. Wei Zhang and Zi Wang have the equal contributions to this study).

16[30] Wen J, Zhang Z, Xu Y, Zhang B, Fei L, Liu H. Unified embedding alignment with missing views inferring for incomplete multi-view clustering. Proceedings of the AAAI conference on artificial intelligence2019. p. 5393-400.

[31] Liu X, Zhu X, Li M, Wang L, Tang C, Yin J, et al. Late fusion incomplete multi-view clustering. IEEE transactions on pattern analysis and machine intelligence. 2018;41:2410-23.

[32] Li Z, Tang C, Zheng X, Liu X, Zhang W, Zhu E. High-order correlation preserved incomplete multi-view subspace clustering. IEEE Transactions on Image Processing. 2022;31:2067-80.

[33] Yuan L, Wang Y, Thompson PM, Narayan VA, Ye J, Initiative AsDN. Multi-source feature learning for joint analysis of incomplete multiple heterogeneous neuroimaging data. NeuroImage. 2012;61:622-32.

[34] Xiang S, Yuan L, Fan W, Wang Y, Thompson PM, Ye J, et al. Bi-level multi-source learning for heterogeneous block-wise missing data. NeuroImage. 2014;102:192-206.

[35] Liu M, Zhang J, Yap P-T, Shen D. View-aligned hypergraph learning for Alzheimer's disease diagnosis with incomplete multi-modality data. Medical image analysis. 2017;36:123-34.

[36] Thung K-H, Yap P-T, Shen D. Multi-stage diagnosis of Alzheimer's disease with incomplete multimodal data via multi-task deep learning. Deep Learning in Medical Image Analysis and Multimodal Learning for Clinical Decision Support: Third International Workshop, DLMIA 2017, and 7th International Workshop, ML-CDS 2017, Held in Conjunction with MICCAI 2017, Québec City, QC, Canada, September 14, Proceedings: Springer; 2017. p. 160-8.

[37] Zhang W, Deng Z, Zhang T, Choi K-S, Wang J, Wang S. Incomplete Multiple View Fuzzy Inference System With Missing View Imputation and Cooperative Learning. IEEE Transactions on Fuzzy Systems. 2021;30:3038-51.

[38] Zhang W, Deng Z, Choi K-S, Wang J, Wang S. Dual Representation Learning for One-Step Clustering of Multi-View Data. arXiv preprint arXiv:220814450. 2022.

[39] Goyal P, Ferrara E. Graph embedding techniques, applications, and performance: A survey. Knowledge-Based Systems. 2018;151:78-94.

[40] Lee DD, Seung HS. Learning the parts of objects by non-negative matrix factorization. Nature. 1999;401:788-91.

[41] Jiang Y, Deng Z, Chung F-L, Wang S. Realizing two-view TSK fuzzy classification system by using collaborative learning. IEEE transactions on systems, man, and cybernetics: systems. 2016;47:145-60.

[42] Friedman JH. On bias, variance, 0/1—loss, and the curse-of-dimensionality. Data mining and knowledge discovery. 1997;1:55-77.